\documentclass{article}
\usepackage{graphicx} 

\usepackage[numbers]{natbib}

\usepackage{hyperref}
\usepackage{url}
\usepackage{graphicx}
\usepackage{placeins}

\usepackage{blindtext}
\usepackage{multicol}

\title{
InhibiDistilbert:
Knowledge Distillation for a ReLU and Addition-based Transformer
}
\author{Tony Zhang and Rickard Brannvall}
\date{December 2024}

\begin{document}
\maketitle

\begin{abstract}
This work explores optimizing transformer-based language models by integrating model compression techniques with inhibitor attention, a novel alternative attention mechanism. Inhibitor attention employs Manhattan distances and ReLU activations instead of the matrix multiplications and softmax activation of the conventional scaled dot-product attention. This shift offers potential computational and energy savings while maintaining model effectiveness. We propose further adjustments to improve the inhibitor mechanism's training efficiency and evaluate its performance on the DistilBERT architecture. 
Our knowledge distillation experiments indicate that the modified inhibitor transformer model can achieve competitive performance on standard NLP benchmarks, including General Language Understanding Evaluation (GLUE) and sentiment analysis tasks.
\end{abstract}

\section*{Introduction.}
Transformer-based language models have revolutionized natural language processing (NLP), achieving state-of-the-art performance across a wide range of tasks, from machine translation to sentiment analysis \citep{vaswani2023attentionneed}. However, the computational and energy demands of these models, particularly those arising from the self-attention mechanism, pose significant challenges for deployment in resource-constrained environments. 
Although highly effective, the self-attention mechanism relies heavily on matrix multiplications, which are computationally expensive and energy-intensive
As the scale of transformer models continues to grow, so does their environmental impact, with studies estimating that training a single large model can emit as much carbon as five cars over their lifetimes \citep{strubell-etal-2019-energy}. This has spurred research into more efficient alternatives, including model compression techniques such as knowledge distillation \citep{sanh2019distilbert} and alternative attention mechanisms, like ReLUFormer \citep{shen2023studyrelusoftmaxtransformer} or Linformer \citep{wang2020linformerselfattentionlinearcomplexity}.
Another alternative is the inhibitor attention \citep{brannvall2023inhibitor}, which was introduced as a means to avoid using the softmax function and matrix multiplications. 

The motivation for this work is driven by the potential advantages of inhibitor attention over conventional dot-product-based attention under low-bit precision quantization. Scaled dot-product attention relies on floating-point matrix multiplication and Softmax activations, which can become challenging when quantized, leading to precision loss. This work takes a first step towards inhibitor transformer model compression by demonstrating that it can be trained via knowledge distillation to perform well on NLP benchmark tasks. Conducted as a Master Thesis project during the fall of 2024, it faced several resource limitations (e.g., access to powerful GPUs). Therefore, it is presented here as a short paper workshop contribution to invite collaborators.

\begin{table}[t]
    \centering
    \caption{Experiment comparing a pre-trained conventional DistilBert with the Inhibitor alternative pre-trained by task-agnostic knowledge distillation. Each model was fine-tuned to the GLUE Benchmarks and the IMDB tasks. The performance on each test was averaged over three runs. }
    \small
    \vspace{0.5cm}
        \begin{tabular}
        {lcc}
            \hline
            \textbf{Models} & 
            \textbf{GLUE} & 
            \textbf{IMDb} \\
            \hline
            Conv. DistilBERT & 77.0 & 
            92.82 \\
            Inhibi.DistilBERT & 74.5 & 
            92.81\\
            \hline
        \end{tabular}
    \label{tab:model-comparison}
\end{table}

\section*{Method}

The dot-product attention score of the conventional Transformer is replaced with the Manhattan distance according to
\begin{equation}
S = \frac{QK^T}{\sqrt{d}} 
\quad
\longrightarrow
\quad
Z_{ij} = \sum_k  \frac{\gamma}{\sqrt{d}} |Q_{ik} - K_{jk}| 
\end{equation}
where $Q, K, V$ are the query, key, and value matrices of \citep{vaswani2023attentionneed} and $d$ is the size of the latent dimension. The attention head output is then similarly replaced
\begin{equation}
H = \mathrm{softmax}(S) V 
\quad
\longrightarrow
\quad
H_{il}' 
= \eta \sum_j \left( ( V_{jl}^+ - \bar{Z}_{ij})^+
+ (V_{jl}^- + \bar{Z}_{ij})^-
\right)
\end{equation}
written with in the notation $(x)^+ = \max(x, 0)$ and $(x)^- = \min(x, 0)$ for the positive and negative ReLU functions, respectrively.  

The attenuating effect for the Inhibitor is obtained by ReLU instead of conventional Softmax. Note how the attention score is applied separately to the positive and negative parts of V, thresholded by the inhibitor attentions score. 

To allow for further calibration of the inhibition effect, a shift is applied to the inhibition score by first centering the score and then adjusting it with a shifting parameter $\delta$
\begin{equation}
\bar{Z}_{ij} = (Z_{ij} - \langle Z_{ij} \rangle_j - \delta)^+
\end{equation}
where $\langle Z_{ij} \rangle_j$ denotes that the mean is calculated over the axis corresponding to index $j$.  The purpose is to control when values from $V$ can pass through unmodified. 
Compared to the previous formulation of the Inhibitor Transformer \cite{brannvall2023inhibitor}, this article introduces a new set of learnable scalar parameters for the inhibitor, 
$\gamma$, $\eta$, and $\delta$, which are specific to each attention head.

\section*{Experiments}

Our experiments were based on the DistilBERT \citep{sanh2019distilbert} paper, 
incorporating elements also from the TinyBERT \citep{jiao-etal-2020-tinybert} and Greedy Layer-Wise Training \citep{bengio2007greedy} papers. 
We conducted two experiments: 1) task-agnostic KD, and 2) task-specific KD. Each is described in more detail below. 
For the task-agnostic KD experiments, instead of using a full-sized BERT model as the teacher as in \citep{sanh2019distilbert}, we used the smaller pre-trained DistilBERT model for computational convenience and simpler alignment.
For the task-specific KD we used a BERT already fine-tuned to the GLUE task was used as teacher. In both cases, weights were initialized from the teacher model.

\paragraph{Experiment 1.}
We transfer knowledge from a dot-product-based DistilBERT teacher model to an inhibitor-based DistilBERT student model using a \textbf{task-agnostic knowledge distillation strategy}:
\begin{itemize}
\item \textbf{Layerwise Training}: Each layer was trained sequentially using Mean Squared Error (MSE) loss to align self-attention outputs. Only the query, key, and value matrices of the current layer were updated, while all other weights remained frozen. The layerwise training followed a bottom-up approach, freezing all layers except for the current one being trained.
\item \textbf{Full-Layer Training}: After layerwise alignment, all layers were unfrozen and trained together using MSE loss applied to hidden states to refine the representations.
\end{itemize}

The initial phase involved layerwise training to align the contextual representations between the teacher and student models using 10\% of the Wikitext-103 corpus. In this phase, all weights in the student model were frozen except for the weight matrices of the query, key, and value components in the current layer being trained. The Mean Squared Error (MSE) loss function was applied to align the context outputs of corresponding layers in both models. Each layer was trained iteratively from the bottom to the top layer (layer 0 to layer 5). After training a layer for two epochs, its weights were frozen, and the next layer in the sequence was unfrozen.

Following the layerwise training, a full-layer training phase was conducted using 60\% of the Wikipedia 20231101 corpus. In this phase, all layers in the student model were unfrozen, allowing parameter updates across the entire network. MSE loss was applied to the hidden states to align the hidden layer outputs between the teacher and student models.

Once the task-agnostic knowledge distillation was completed, the final weights of the inhibitor DistilBERT model were stored and used as the foundation for fine-tuning to more specific NLP tasks. 

\paragraph{Experiment 2.}
We performed \textbf{task-specific knowledge distillation} using a fine-tuned BERT model on GLUE benchmark \citep{wang2019gluemultitaskbenchmarkanalysis} tasks. The goal was to transfer task-specific knowledge to the inhibitor model by using the BERT model as a teacher. The loss components consisted of:
\begin{itemize}
    \item \textbf{Soft Probability Distillation Loss}: The distillation loss function uses the teacher model's soft probabilities to encourage the student model to replicate the teacher's predictions.
    \item \textbf{Hidden State Loss}: To help guide the inner layers of the student model toward better alignment with the teacher's representations.
\end{itemize}

In the second experiment, we extended our initial task-agnostic distillation approach by implementing task-specific knowledge distillation with a fine-tuned BERT model on GLUE benchmark tasks. The objective was to transfer the specialized knowledge from the BERT model (acting as the teacher) to the inhibitor model (acting as the student). This process involved two main components: the Soft Probability Distillation Loss, which used the teacher model's soft probabilities to guide the student model in mimicking the teacher's predictions, and the Hidden State Loss, which aimed to align the inner layers of the student model more closely with the teacher's internal representations.

\begin{table}[b]
    \centering
    \caption{Supplementary results showing that the performance on GLUE is somewhat weaker for Task-Specific Knowledge Distillation (KD, bottom row) compared to the case with Fine-Tuning after Task-Agnostic Distillation (FT, same as Table \ref{tab:model-comparison}). Task scores are averaged over three runs.\\}
    \resizebox{\textwidth}{!}{
        \begin{tabular}{l|rrrrrrrrr}
            \hline
            \textbf{Models} 
            & \textbf{CoLA} & \textbf{MNLI} & \textbf{MRPC} & \textbf{QNLI} & \textbf{QQP} & \textbf{RTE} & \textbf{SST-2} & \textbf{STS-B} & \textbf{WNLI} \\
            \hline
            Conv. DistilBERT (FT) 
            & 51.3 & 82.2 & 87.5 & 89.2 & 88.5 & 59.9 & 91.3 & 86.9 & 56.3 \\
            Inhibi.DistilBERT (FT) 
            & 40.0 & 79.2 & 86.8 & 85.4 & 89.5 & 59.2 & 90.2 & 83.5 & 56.3 \\
            Inhibi.DistilBERT (KD) 
            & 47.5 & 72.2 & 77.0 & 80.0 & 63.4 & 47.3 & 91.0 & 83.5 & 56.3 \\
            \hline
        \end{tabular}
    }
    \label{tab:model-comparison2}
\end{table}

\section*{Results}
We evaluate our inhibitor-based DistilBERT model against the conventional scaled dot-product attention DistilBERT on the GLUE benchmark \citep{wang2019gluemultitaskbenchmarkanalysis}, which consists of nine different language understanding tasks.
We fine-tuned each model for three epochs using the AdamW optimizer in accordance with standard practices also followed in the original DistilBERT paper. 

The performance comparison in Table \ref{tab:model-comparison} indicates that a fine-tuned inhibitor DistilBERT achieves competitive accuracy, with a modest 3.2\% average drop on GLUE compared to dot-product DistilBERT across the different tasks. 
For reference, results for the IMDB sentiment analysis task are also presented, which show no distinguishable difference in the performance on this simple task. 

Looking at the detailed performance on each separate GLUE task in Table \ref{tab:model-comparison2} indicates that the fine-tuned task agnostic inhibitor DistilBERT (FT Inhibi.DistilBERT) achieves competitive accuracy across most tasks but faced a notable challenge for the CoLA task. This would require a more detailed analysis as to why.
We note that a performance drop may be expected as we used the original Distilbert model both as a teacher model and as a benchmark baseline. 

Task-specific knowledge distillation (KD Inhibi.DistilBERT) lags behind, indicating that improvements in layer alignment and training strategies are needed. Although the task-specific model performs somewhat better on the CoLA benchmark, 
it shows materially worse results for most other benchmarks, particularly MNLI, MRPC, QNLI, and QQP.
We note that the training data for the GLUE tasks is sparse and that we should not expect strong results. 

Results from a third experiment on computational efficiency were inconclusive (details can be provided upon inquiry). While our theoretical analysis suggested potential energy savings, the practical experiments showed higher energy consumption and lower throughput on the available computer architecture, indicating the need for further experimentation on more specialized hardware.

\section*{Conclusions}

We demonstrate that an inhibitor-based Transformer trained by knowledge distillation can achieve competitive performance on NLP benchmarks while relying on simpler arithmetic operations. However, while theoretical analysis suggests potential energy savings, actual measurements on conventional server hardware showed higher energy consumption and lower throughput compared to traditional dot-product attention. This discrepancy underscores the need for specialized hardware, such as custom FPGA designs optimized for ReLU and addition-based operations, to fully realize the theoretical benefits of this novel attention mechanism. The original report can be provided upon inquiry for a detailed analysis of energy efficiency and throughput.

\textbf{Limitations.}
Several limitations must be acknowledged. First, the generalizability of the inhibitor attention mechanism to other transformer architectures, such as GPT or Vision Transformers, remains unexplored. 
The scalability of the inhibitor attention mechanism for very large-scale models was also not addressed.  
A more comprehensive comparison with other efficient attention mechanisms, such as Linformer or Performer, would provide a clearer understanding of its relative advantages and disadvantages. 
Testing was limited to IMDB and GLUE benchmarks but should include more modern, harder, or multilingual tasks like SuperGLUE and XTREME.
Additionally, the impact of low-bit precision quantization on the inhibitor attention mechanism was not extensively studied, which may affect the model's performance. 
Furthermore, the practical challenges and performance of deploying inhibitor-based models in real-world applications, particularly in resource-constrained environments, were not examined.
Lastly, the energy consumption analysis was limited to conventional server hardware, necessitating further investigation using specialized hardware.

\textbf{Future Work. }
We will focus on further optimizations, including specialized hardware implementations and model compression by quantization-aware training. Additionally, exploring knowledge transfer from a full-sized BERT model instead of DistilBERT should improve performance. With more powerful GPUs, direct pretraining of Inhibitor Transformers and experiments with larger, modern architectures would be feasible. Additionally, we would like to assess the mechanism's performance in generative language models (GPT family) and Vision Transformers. Specifically, we plan to investigate the applicability and performance of inhibitor attention across a broader range of tasks and modalities. Further research will also include deployment studies to assess the feasibility and benefits of inhibitor-based models in practical scenarios.

\bibliography{main}
\bibliographystyle{plain}


\begin{table}[b]
\parbox{.45\linewidth}{
    \centering
    \caption{Hyperparameters for the layer-wise training.\\}
    \begin{tabular}{lr}
        \textbf{Hyperparameter} & \textbf{Value} \\
        Number of Layers & 6 \\
        Hidden size & 768 \\
        FFN inner hidden size & 3072 \\
        Attention heads & 12 \\
        Attention head size & 64 \\
        Dropout & 0.1 \\
        Attention Dropout & 0.1 \\
        Warmup Ratio & 5\% \\
        Peak Learning Rate & 5e-4 \\
        Batch Size & 16 \\
        Gradient accumulation steps & 4 \\
        Epochs & 2 \\
        Learning Rate Decay & Cosine \\
        Adam $\epsilon$ & 1e-8 \\
        Adam $\beta_1$ & 0.9 \\
        Adam $\beta_2$ & 0.999 \\
    \end{tabular}
    \label{tab:hyperparameters1}
}
\parbox{.10\linewidth}{
\quad
}
\parbox{.45\linewidth}{

    \centering
    \caption{Hyperparameters for the full-layer training.\\}
    \begin{tabular}{lr}
        \textbf{Hyperparameter} & \textbf{Value} \\
        Number of Layers & 6 \\
        Hidden size & 768 \\
        FFN inner hidden size & 3072 \\
        Attention heads & 12 \\
        Attention head size & 64 \\
        Dropout & 0.1 \\
        Attention Dropout & 0.1 \\
        Warmup Ratio & 5\% \\
        Peak Learning Rate & 3e-4 \\
        Batch Size & 16 \\
        Gradient accumulation steps & 32 \\
        Epochs & 3 \\
        Learning Rate Decay & Cosine \\
        Adam $\epsilon$ & 1e-8 \\
        Adam $\beta_1$ & 0.9 \\
        Adam $\beta_2$ & 0.999 \\
    \end{tabular}
    \label{tab:hyperparameters2}
}
\end{table}


\begin{table}[b]
\parbox{.45\linewidth}{
    \centering
    \caption{Hyperparameters for the task-specific knowledge distillation.\\}
    \begin{tabular}{lr}
        \textbf{Hyperparameter} & \textbf{Value} \\
        Number of Layers & 6 \\
        Hidden size & 768 \\
        FFN inner hidden size & 3072 \\
        Attention heads & 12 \\
        Attention head size & 64 \\
        Dropout & 0.1 \\
        Attention Dropout & 0.1 \\
        Warmup Ratio & 0 \\
        Peak Learning Rate & 2e-5 \\
        Batch Size & 16 \\
        Gradient accumulation steps & 0 \\
        Epochs & 3 \\
        Learning Rate Decay & Linear\\
        Adam $\epsilon$ & 1e-8 \\
        Adam $\beta_1$ & 0.9 \\
        Adam $\beta_2$ & 0.999 \\
        Temperature & 4 \\
        Distillation loss weight & 0.5 \\
        Hidden state loss weight & 0.5 \\
    \end{tabular}
    \label{tab:hyperparameters3}
}
\parbox{.10\linewidth}{
\quad
}
\parbox{.45\linewidth}{
    \centering
    \caption{Hyperparameters for fine-tuning after task-agnostic distillation.\\}
    \begin{tabular}{lr}
        \textbf{Hyperparameter} & \textbf{Value} \\
        Number of Layers & 6 \\
        Hidden size & 768 \\
        FFN inner hidden size & 3072 \\
        Attention heads & 12 \\
        Attention head size & 64 \\
        Dropout & 0.1 \\
        Attention Dropout & 0.1 \\
        Warmup Ratio & 0 \\
        Peak Learning Rate & 2e-5 \\
        Batch Size & 16 \\
        Epochs & 3 \\
        Learning Rate Decay & Linear\\
        Adam $\epsilon$ & 1e-8 \\
        Adam $\beta_1$ & 0.9 \\
        Adam $\beta_2$ & 0.999 \\
    \end{tabular}
    \label{tab:hyperparameters4}
}
\end{table}

\end{document}